\documentclass[10pt,twocolumn,letterpaper]{article}

\usepackage[pagenumbers]{cvpr} 

\usepackage{pifont}

\definecolor{cvprblue}{rgb}{0.21,0.49,0.74}
\usepackage[pagebackref,breaklinks,colorlinks,allcolors=cvprblue]{hyperref}

\def\paperID{} 
\def\confName{CVPR}
\def\confYear{2026}

\title{TAUE: Training-free Noise Transplant and Cultivation Diffusion Model}

\author{ Daichi Nagai$^{1}$\thanks{Equal contribution.}, Ryugo Morita$^{2}$\footnotemark[1], Shunsuke Kitada$^{1}$, Hitoshi Iyatomi$^{1}$ \\ $^{1}$Faculty of Science and Engineering, Hosei University, Tokyo, Japan \\ $^{2}$RPTU Kaiserslautern-Landau \& DFKI GmbH, Kaiserslautern, Germany \\ 
}

\begin{document}
\maketitle
\begin{abstract}
Despite the remarkable success of text-to-image diffusion models, their output of a single, flattened image remains a critical bottleneck for professional applications requiring layer-wise control. Existing solutions either rely on fine-tuning with large, inaccessible datasets or are training-free yet limited to generating isolated foreground elements, failing to produce a complete and coherent scene. To address this, we introduce the Training-free Noise Transplantation and Cultivation Diffusion Model (TAUE), a novel framework for layer-wise image generation that requires neither fine-tuning nor additional data.
TAUE embeds global structural information from intermediate denoising latents into the initial noise to preserve spatial coherence, and integrates semantic cues through cross-layer attention sharing to maintain contextual and visual consistency across layers.
Extensive experiments demonstrate that TAUE achieves state-of-the-art performance among training-free methods, delivering image quality comparable to fine-tuned models while improving inter-layer consistency. Moreover, it enables new applications, such as layout-aware editing, multi-object composition, and background replacement, indicating potential for interactive, layer-separated generation systems in real-world creative workflows.
\end{abstract}    
\begin{figure}[t]
    \centering
    \includegraphics[width=0.91\columnwidth]{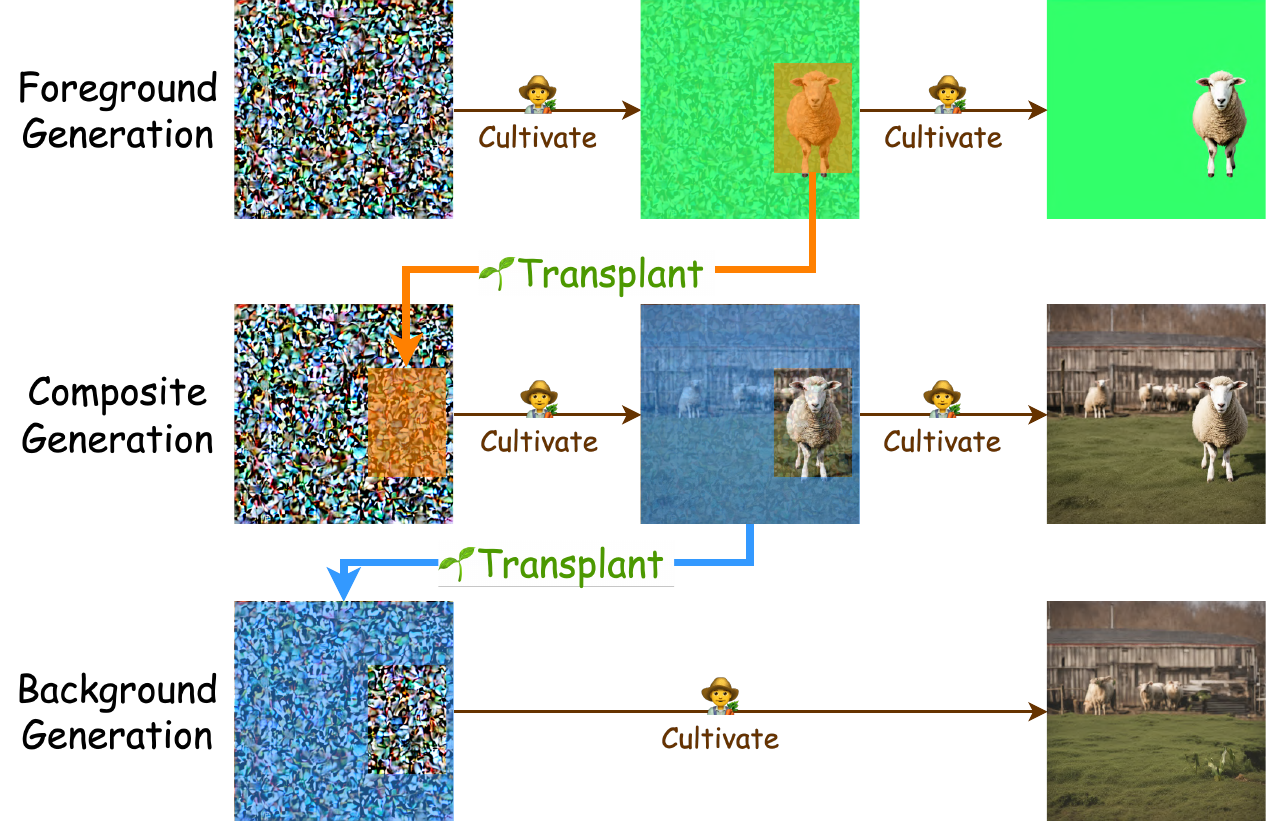} 
    \caption{
    TAUE introduces a training-free method for layer-wise image generation, enabling independent foreground–background synthesis with a coherent composite output.
    Our method further enables applications such as layout and size control, disentangled multi-object generation, and background replacement, broadening the creative and practical capabilities of real-world workflows.
    }
    \label{fig:teaser}
\end{figure}

\section{Introduction}
\label{sec:intro}

Diffusion models have revolutionized creative workflows by enabling the synthesis of photorealistic and intricate images from text~\cite {ho2020denoising, song2020denoising, rombach2022high, hu2024diffumatting}. Yet, this transformative power is constrained by a critical limitation: they typically generate only single-layered, flat images. These flat images severely restrict post-hoc manipulation, as individual elements are inextricably fused. In professional domains such as art, design, and animation, where refining complex compositions is crucial, this lack of layer-wise control presents a significant bottleneck, forcing practitioners to engage in laborious manual segmentation and inpainting.

In response, most existing work has leveraged fine-tuning for layer-wise image generation. These methods attempt to denoise multiple layers simultaneously using masks~\cite{zhang2023text2layer, huang2024layerdiff, fontanella2024generating, kang2025layeringdiff} or employ specialized alpha-channel autoencoders~\cite{zhang2024transparent, dalva2024layerfusion, huang2025psdiffusion, pu2025art}. While effective to some extent, these approaches depend on large-scale, curated datasets and prohibitive training costs. Critically, these datasets are often proprietary or inaccessible, creating a significant barrier to entry that hinders reproducible research and slows community progress.
To circumvent this data dependency, training-free approaches have been explored~\cite{quattrini2024alfie, morita2025tkg, zou2025zero}. However, these methods focus exclusively on generating isolated foregrounds and do not attempt to produce a corresponding background. This inherently limits them to partial solutions, leaving a critical research gap for a framework capable of layer-wise image generation without fine-tuning.

To address this gap, we propose the novel Training-free Noise Transplant and Cultivation Diffusion Model (TAUE)\footnote{\textit{Taue} is the Japanese term for transplanting rice seedlings, symbolizing the nurturing and cultivation process, which parallels our model's core concept of cultivating latent representations.}, independently generates foreground and background, and coherent
composite images. 
Instead of regenerating all layers from scratch, TAUE reuses intermediate latents extracted during the denoising process and transplants them into the initial noise of subsequent generations, enabling structural continuity across layers. 
Moreover, TAUE introduces cross-layer attention sharing, enabling the exchange of semantic information between the foreground and background to maintain scene-level coherence and context alignment. 
As a result, TAUE produces consistent multi-layered outputs and supports complex compositional editing and interactive layer-separated generation without any fine-tuning or additional data.

Extensive experiments demonstrate that TAUE achieves state-of-the-art results among training-free methods, improving layer-wise consistency while matching the quality of fine-tuned models. Our contributions are as follows:

\begin{itemize}
    \item We introduce TAUE, a training-free framework for layer-wise image generation that produces coherent foreground, background, and composite images without any fine-tuning or external datasets.
    \item TAUE combines latent transplantation, which embeds structural information from intermediate latent into the initial noise, with cross-layer attention sharing to propagate semantic cues, achieving consistent and contextually aligned multi-layer generation.
    \item Our method outperforms existing models, delivering results
    that match or surpass fine-tuned alternatives while maintaining lower computational costs.
    \item TAUE enables low-cost, practical, and interactive applications such as complex compositional editing and layer-separated image generation.

\end{itemize}


\begin{figure*}[t]
\centering
\includegraphics[width=0.95\textwidth]{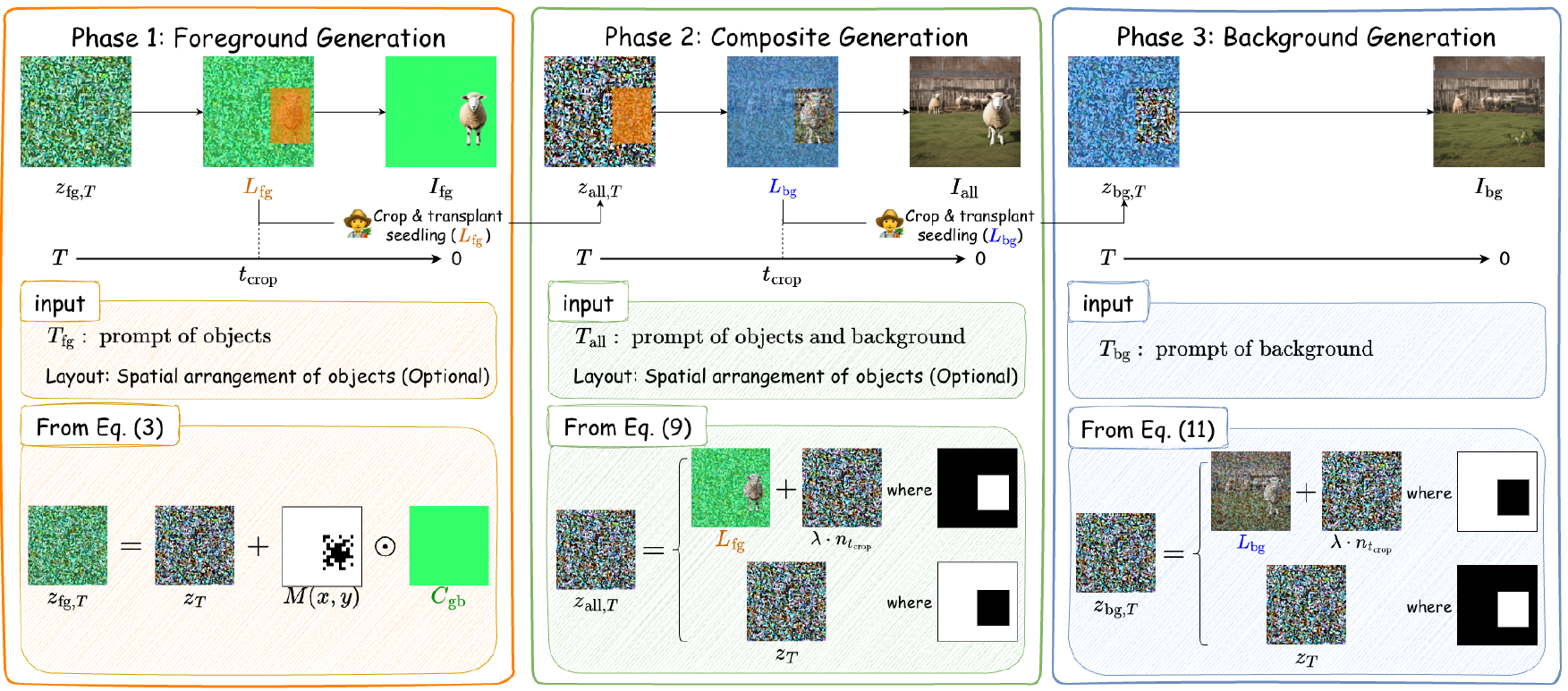} 
\caption{
    TAUE consists of three stages: (1) \textbf{Foreground Generation}, where an object is generated from noise and an intermediate latent is extracted; 
(2) \textbf{Composite Generation}, where the extracted latent is transplanted into a new denoising trajectory to synthesize a coherent foreground–background scene through cross-layer attention sharing; and 
(3) \textbf{Background Generation}, which mirrors the composite phase with the object mask inverted to refine the background region. 
This three-stage design enables TAUE to maintain spatial and semantic coherence across layers, producing unified, multi-layered image generation without any fine-tuning or additional data.
}
\label{fig: model}
\end{figure*}

\section{Related Works}
\label{sec:related}

\subsection{Initial Noise of Diffusion Model}

Diffusion models turn Gaussian noise into structured data through a step-by-step refinement process. Tailoring the initial noise improves image fidelity and alignment, with recent studies focusing on optimal noise selection~\cite{xu2024good, wang2024silent, guo2024initno, eyring2024reno, ahn2024noise} and latent optimization~\cite{chen2024find, zhou2024golden, qi2024not, eyring2025noise}.
Furthermore, identifying optimal noise seeds and focusing on specific noise regions affect image quality and object placement~\cite{ban2024crystal, han2025spatial, izadi2025fine}, while noise manipulation enables advanced tasks for layout control~\cite{shirakawa2024noisecollage, mao2023semantic}, editing~\cite {mao2023guided, chen2024tino}, and video generation~\cite{wu2024freeinit, morita2025tkg}.

Building on these insights, we propose embedding the intermediate latent into the initial noise of the diffusion process. 
This allows the model to preserve the structural information of the object such as its shape and spatial configuration while reconstructing the image with improved layer-wise consistency and generating entirely new content guided by the text condition.

\begin{table}[t]
\caption{Qualitative capability comparison between our TAUE and well-established layer-wise image generation methods such as LayerDiffuse~\cite{zhang2024transparent}, ART~\cite{pu2025art}, Alfie~\cite{quattrini2024alfie}. TAUE uniquely enables training-free generation of complete scenes with background synthesis, multi-object disentanglement, semantic harmonization, and layout controllability surpassing both fine-tuned and training-free baselines.}
\small
\centering
\begin{tabular}{@{}lcccc@{}}
\toprule
               & \multicolumn{2}{c}{Fine-tuned} & \multicolumn{2}{c}{\textbf{Training-free}} \\ \cmidrule(lr){2-3} \cmidrule(r){4-5}
               & LayerDiffuse       & ART                & Alfie              & \textbf{Ours}                  \\ \cmidrule(r){1-1} \cmidrule(lr){2-2} \cmidrule(lr){3-3} \cmidrule(lr){4-4} \cmidrule(l){5-5}
Background     & \ding{51}       & \ding{51}       & \ding{55}           & \ding{51}          \\
Multi-object   & \ding{51}       & \ding{51}       & \ding{55}           & \ding{51}          \\
Harmonization  & \ding{55}           & \ding{55}           & \ding{55}           & \ding{51}          \\
Layout specification & \ding{55}           & \ding{51}       & \ding{55}           & \ding{51}          \\ \bottomrule
\end{tabular}

\label{tab:hikaku}
\end{table}

\subsection{Layer-wise Image Generation}

Layer-wise image generation is essential for professional applications that require fine-grained compositional control and editing.~\cite{dai2025trans, morita2023interactive, burgert2024magick}
Fine-tuned approaches typically achieve this using special masks~\cite{zhang2023text2layer, huang2024layerdiff, fontanella2024generating, kang2025layeringdiff} or alpha-channel autoencoders~\cite{zhang2024transparent, dalva2024layerfusion, huang2025psdiffusion, pu2025art}. However, these methods rely heavily on large-scale, proprietary datasets, which limits their practicality and reproducibility. 

Training-free alternatives~\cite{morita2025tkg, quattrini2024alfie, zou2025zero} eliminate needs for data collection; however, they fail to produce complete scenes, generating only isolated foregrounds. In contrast, our approach is the first zero-shot, complete layer-wise image generation without fine-tuning or external data.

\subsection{Layer Decomposition and Composition}

An alternative to direct layered generation is post-hoc decomposition, where a flat image is broken into distinct layers. This research focuses on building large-scale datasets to train decomposition models. These datasets may offer either fine-grained part masks, which lack the transparency needed for flexible compositing~\cite{liu2024object}, or full RGBA layers~\cite{huang2025dreamlayer}. However, such datasets are often curated by other models and often fail to capture complex visual effects like soft shadows and reflections, limiting their utility for seamless editing~\cite{burgert2024magick, tudosiu2024mulan, yang2025generative}.

Moreover, beyond decomposition, harmonizing extracted layers into a cohesive image poses a distinct challenge. This step typically requires costly fine-tuning to blend lighting and color, a process that can alter the original appearance of elements~\cite{winter2024objectdrop, zhang2020deep, huang2025psdiffusion}. These dual challenges, dependence on large datasets and expensive harmonization, underscore fundamental limitations of this pipeline. In contrast, TAUE bypasses these issues by generating coherent foreground, background, and composite layers simultaneously in zero-shot fashion, ensuring consistency from the outset without datasets or post-hoc correction.
\section{TAUE}
\subsection{Overall}
TAUE utilizes Latent Diffusion Model (LDM)~\cite{rombach2022high} to achieve layer-wise image generation without requiring fine-tuning or additional datasets. 
As shown in Fig.~\ref{fig: model}, TAUE is divided into three phases: (1) Foreground Generation, (2) Composite Generation, and (3) Background Generation. 
We use three distinct text prompts --- $T_{\text{fg}}$, $T_{\text{bg}}$, and $T_{\text{all}}$ --- for foreground, background, and composite generation.
\begin{enumerate}[label=(\arabic*), leftmargin=*]
    \item We generate the foreground object on a uniform background to obtain a clean foreground layer $I_{\text{fg}}$ (\S\ref{sec: method_fore}). During this step, an intermediate foreground latent $L_{\text{fg}}$ is extracted, encoding structural features of the object.
    \item $L_{\text{fg}}$ is transplanted into the initial noise $z_{\text{all}, T}$ for composite generation, yielding both the intermediate background latent $L_{\text{bg}}$ and the composite scene $I_{\text{all}}$ (\S\ref{sec: method_comb}).
    \item $L_{\text{bg}}$ is transplanted into the initial noise $z_{\text{bg}, T}$ for generating the background $I_{\text{bg}}$, ensuring consistency between the foreground, background, and composite scene (\S\ref{sec: method_back}).
\end{enumerate}
The final output comprises three distinct layers: the foreground $I_{\text{fg}}$, background $I_{\text{bg}}$, and composite scene $I_{\text{all}}$, forming a coherent image.

\subsection{Foreground Generation}
\label{sec: method_fore}
TAUE first generates an isolated foreground object according to the foreground prompt $T_{\text{fg}}$.
Inspired by TKG-DM~\cite{morita2025tkg}, which demonstrated that adding channel-wise biases to the initial Gaussian noise $z_T \in \mathbb{R}^{4\times H/8\times W/8}$ enables color-controlled image synthesis, 
we apply a similar strategy to guide the background color in the latent space. 
Specifically, we blend a green background latent vector $C_{\text{gb}} = [0,1,1,0]$ ($C_{\text{gb}} \in \mathbb{R}^{4\times H/8\times W/8}$) into the initial noise $z_T$ within a spatial mask $M$:
\begin{equation}
    z_{\text{fg}, T} = (1 - M) \odot z_T + M \odot ((1 - \alpha) z_T + \alpha C_{\text{gb}}),
\end{equation}
where $\alpha$ controls the blending strength of the background color. 
The mask $M$ defines the spatial region for foreground synthesis and is detailed in the following subsection (\S\ref{sec:layout_spec}). 
The resulting latent $z_{\text{fg}, T}$ is then denoised using the foreground prompt $T_{\text{fg}}$ to produce the isolated foreground image $I_{\text{fg}}$.

\subsubsection{Layout Specification}
\label{sec:layout_spec}
Existing methods typically employ a Gaussian or binary box-shaped mask to localize the foreground region as a spatial mask $M$. 
However, this approach often produces artifacts along the mask boundary, as the generated content tends to conform too closely to the mask shapes. 
To overcome this limitation, we redefine $M$ as a \emph{probabilistic layout mask} that decouples object generation from mask edges by introducing a spatially weighted sampling scheme.

Given a bounding box centered at $(o_x, o_y)$ with width $w$ and height $h$, we define a radially symmetric Gaussian distribution confined to this region:

\begin{equation}
P(x,y)=
\begin{cases}
\exp\!\left(
-\frac{1}{2\sigma^2}
\left[
\left(\frac{x-o_x}{w/2}\right)^2+
\left(\frac{y-o_y}{h/2}\right)^2
\right]
\right),\\
\quad \text{if } |x-o_x|\le w/2,\ |y-o_y|\le h/2,\\
0,\quad \text{otherwise}.
\end{cases}
\end{equation}
where $P(x, y)$ serves as an \emph{object retention score}, indicating how strongly the original noise at each pixel is preserved. 
Larger values promote stable synthesis near the center, while lower values allow flexible adaptation near edges. 
After rescaling $P(x,y)$ to $[p_{\min}, p_{\max}]$, we generate a binary mask by stochastic comparison with a random matrix $R(x,y)$:
\begin{equation}
    M(x,y) =
    \begin{cases}
        1 & \text{if } R(x,y) > P(x,y), \\
        0 & \text{otherwise.}
    \end{cases}
\end{equation}
This probabilistic masking allows smooth transitions at object boundaries, solving artifacts tied to the mask contour and enabling flexible control over position and scale.
The resulting mask $M$ is then used for latent blending during foreground generation (\S\ref{sec: method_fore}), ensuring that the generated foreground follows the intended layout while remaining structurally coherent.

\subsubsection{Intermediate Latent Extraction}
During denoising, we cache the intermediate latent tensor at a specific timestep $t_{\text{crop}}$, corresponding to a predefined denoising ratio $r_{\text{crop}}$. 
Formally,
\begin{equation}
    L_{\text{fg}} = z_{\text{fg}, t_{\text{crop}}} \in \mathbb{R}^{4\times H/8\times W/8},
    \label{eq:Lfg-def}
\end{equation}
where $t_{\text{crop}} = \lfloor T \cdot (1 - r_{\text{crop}}) \rfloor$. 
We refer to this cached latent, which encodes the geometric and semantic structure of the foreground object and will be transplanted in subsequent stages to guide composite and background generations.

\begin{figure}[t]
\centering
\includegraphics[width=0.95\columnwidth]{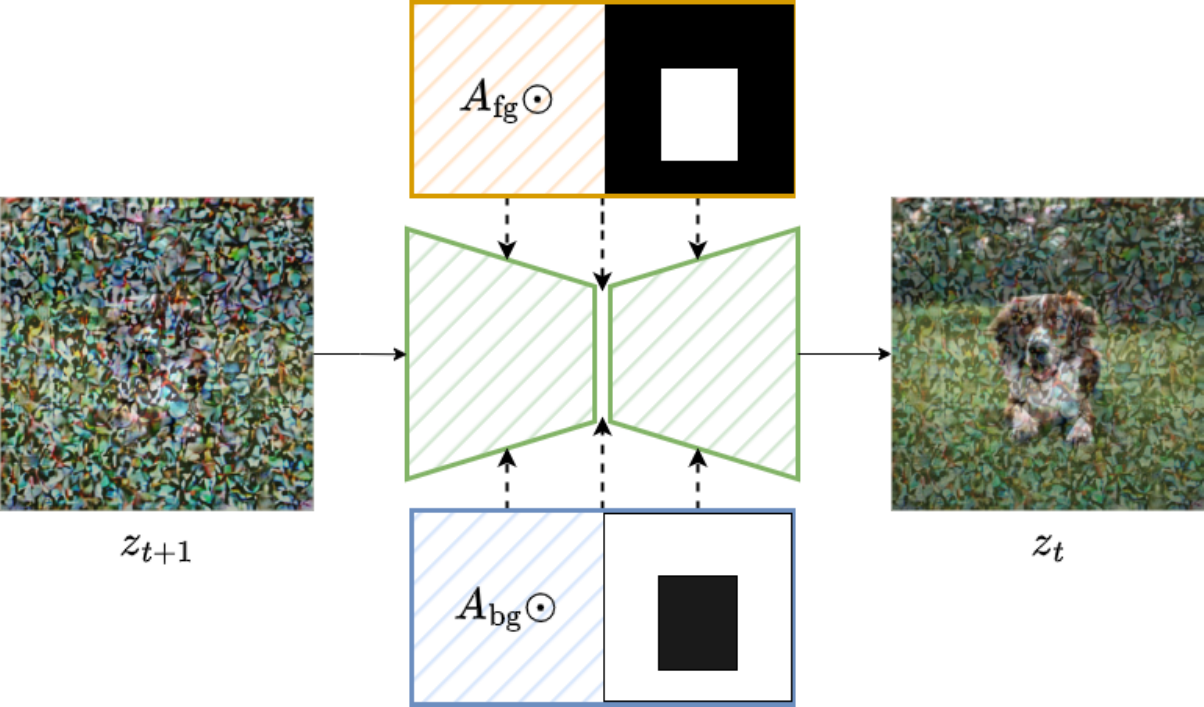}
\caption{
    Illustration of the cross-attention blending mechanism. The foreground prompt is applied to object regions $m_{\text{obj}}$, while the background prompt is applied to non-object regions $1-m_{\text{obj}}$. This enables precise control over foreground-background composition, ensuring cohesive integration of both layers in the final composite scene.
}
\label{fig: attention}
\end{figure}

\begin{figure*}[t]
\centering
\includegraphics[width=0.95\textwidth]{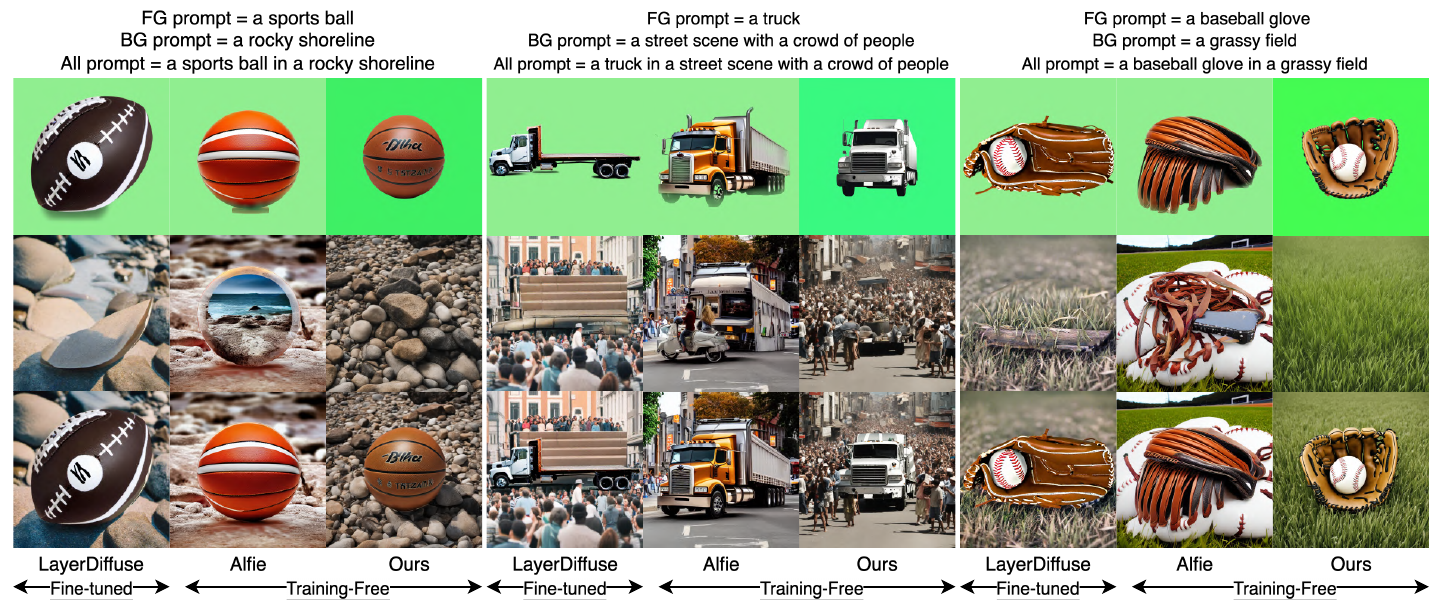} 
\caption{
Qualitative comparison of layer-wise image generation. For each case, we show the foreground, background, and composite image generated by LayerDiffuse~\cite{zhang2024transparent}, Alfie~\cite{quattrini2024alfie}, and our method.  TAUE consistently produces spatially aligned and semantically coherent multi-layer outputs, achieving realistic integration of foreground and background without fine-tuning or inpainting.
}
\label{fig: result}
\end{figure*}

\subsection{Composite Generation}
\label{sec: method_comb}

Just as a seedling can shape the ecosystem it grows in, transplanting latent noise from foreground into new generative processes can guide the formation of coherent visual scenes.  
In this phase, we generate a coherent composite scene by transplanting the intermediate latent $L_{\text{fg}}$ from the foreground phase and cultivating it within the diffusion process. This is guided by an object mask and reinforced through spatially-aware attention and noise control mechanisms.

\subsubsection{Object Region Mask}
We begin by localizing the object region in the latent space using two complementary signals: latent channel activations and cross-attention maps. 
During foreground generation, a green latent vector $C_{\text{gb}}$ is injected into the background region of the initial latent, elevating activations in channels $c = 1$ and $c = 2$. 
The object region, unaffected by this injection, exhibits low activation in these channels. 
We construct a smooth activation map as:
\begin{equation}
    v_{\text{gb}}(x,y) = \mathcal{G}_{\sigma}(L_{\text{fg}}^{(1)} + L_{\text{fg}}^{(2)}),
\end{equation}
where $\mathcal{G}_{\sigma}$ is Gaussian blur operator with parameter $\sigma$, and $L_{\text{fg}}^{(1)}$ and $L_{\text{fg}}^{(2)}$ denote the 1st and 2nd channels of the seedling latent $L_{\text{fg}}$ after foreground generation.
In parallel, cross-attention maps derived from the foreground prompt $T_{\text{fg}}$ highlight semantically relevant spatial regions. Combining these cues, we define the binary object mask $m_{\text{obj}}$ as:
\begin{equation}
    m_{\text{obj}}(x,y)=
\mathbf{1}\bigl[\,v_{\text{gb}}(x,y)<\tau_{\text{bg}}\;
\land\;
A_{\text{fg}}(x,y)>\tau_{A}\bigr],\label{eq:mobj}
\end{equation}
where $\tau_{\text{bg}}$ and $\tau_{\text{attn}}$ are predefined thresholds. This mask ensures accurate spatial localization of the foreground object.
Here, $A_{\text{fg}} \in [0,1]^{H/8\times W/8}$ denotes the token-aggregated cross-attention map of the foreground prompt $T_\text{fg}$.
For every latent location $(x,y)$, we set the object mask $m_{\text{obj}}(x,y)$ to~1 if (i) the green-background response $v_{\text{gb}}(x,y)$ falls below a background threshold $\tau_{\text{bg}}$, and (ii) the attention weight $A_{\text{fg}}(x,y)$ exceeds an attention threshold $\tau_{A}$.
This conjunctive criterion retains only spatial positions that are simultaneously \emph{not} dominated by the injected green background and \emph{strongly attended} by the foreground textual tokens.

\subsubsection{Cross-Attention Shearing} 
To achieve semantic coherence between the foreground and background, we modulate the cross-attention layers using the object mask $m_{\text{obj}}$. As shown in Fig.~\ref{fig: attention}, the foreground prompt $T_\text{fg}$ is applied only to the object region, and the background prompt $T_\text{fg}$ to the rest. 
We obtain the mixed cross-attention tensor $A_{\text{mix}}\in\mathbb{R}^{H/8\times W/8\times d}$ by a pixel-wise convex combination of the foreground‑conditioned attention $A_{\text{fg}}$ and the background-conditioned attention $A_{\text{bg}}$:
\begin{equation}
    A_{\text{mix}}
      = m_{\text{obj}}\odot A_{\text{fg}}
      + (1-m_{\text{obj}})\odot A_{\text{bg}},
\end{equation}
where $m_{\text{obj}}\in\{0,1\}^{H/8\times W/8}$ is the binary object mask based on the object's spatial layout. The mask is broadcast over the $d$ attention channels, ensuring that foreground tokens dominate inside the object region while background tokens govern elsewhere. 

\subsubsection{Noise Transplant and Cultivation}
We initialize the composite generation with a blended latent that preserves object details while introducing new background content:
\begin{equation}
    z'_{\text{all}, T} = m_{\text{obj}} \odot (L_{\text{fg}} + \lambda \cdot n_{t_{\text{crop}}}) + (1 - m_{\text{obj}}) \odot z_T,
\end{equation}
where $L_{\text{fg}}$ is the foreground seedling latent, $n_{t_{\text{crop}}}$ is the predicted noise at the crop timestep, $z_T \sim \mathcal{N}(0,1)$, and $\lambda$ controls noise intensity.
To further enhance spatial details, we apply a Laplacian filter to the seedling latent:
\begin{equation}
    z_{\text{all}, T} = m_{\text{obj}} \cdot (f(L_{\text{fg}}) + \lambda \cdot n_{t_{\text{crop}}}) + (1 - m_{\text{obj}}) \cdot z_T,
\end{equation}
where $f(\cdot)$ is a high-pass filter. Denoising starts from $z_{\text{all}, T}$, with noise blended at each timestep as follows:
\begin{equation}
n_t =
\begin{cases}
    m_{\text{obj}} \odot n_{t_{\text{crop}}} + (1 - m_{\text{obj}}) \odot n_t & \text{if } t_{\text{crop}} \leq t, \\
    n_t & \text{otherwise}.
\end{cases}
\end{equation}
This two-stage scheme fixes the foreground while allowing the background to evolve, ensuring semantic alignment and visual coherence.
The composite image $I_{\text{all}}$ is obtained at the final step, and the intermediate latent $L_{\text{bg}}$ at timestep $t_{\text{crop}}$ is passed to the following background generation phase.


\subsection{Background Generation}
\label{sec: method_back}
This phase mirrors the composite generation but with the object mask inverted. 
Specifically, we transplant the background latent $L_{\text{bg}}$ into the complementary region $(1 - m_{\text{obj}})$ and apply the same noise control scheme symmetrically to cultivate a coherent background $I_{\text{bg}}$ consistent with the composite layers.
Unlike the composite phase, we release the mask constraint in the attention layers and apply the background cross-attention $A_{\text{bg}}$ to all spatial positions, allowing the background prompt to globally refine lighting, color, and contextual harmony across the entire scene.

\begin{table*}[t]
\caption{
    Quantitative comparison of overall image quality and layer-wise reconstruction performance. 
    We report FID, CLIP-I, and CLIP-S for holistic quality, and PSNR, SSIM, and LPIPS for foreground (fg) and background (bg) reconstruction. 
    Lower is better for FID and LPIPS ($\downarrow$), and higher is better for CLIP, PSNR, and SSIM ($\uparrow$). 
    The best and second-best results are shown in bold and underlined, respectively.
    TAUE outperforms both fine-tuned and training-free baselines across most metrics, with its layout-aware variant achieving the best foreground fidelity and overall quality.
}
\small
\centering
\begin{tabular}{@{}lrrrrrrrrr@{}}
\toprule
\multicolumn{1}{c}{} & \multicolumn{3}{c}{Overall Quality}                                                                                  & \multicolumn{6}{c}{Layer-wise Reconstruction Quality}                                                                                                                                                                                                                                               \\ \cmidrule(r){2-4} \cmidrule(l){5-10}
\multicolumn{1}{c}{} & \multicolumn{1}{c}{FID $\downarrow$} & \multicolumn{1}{c}{CLIP-I $\uparrow$} & \multicolumn{1}{c}{CLIP-S $\uparrow$} & \multicolumn{1}{c}{PSNR$_\text{fg} \uparrow$} & \multicolumn{1}{c}{PSNR$_\text{bg} \uparrow$} & \multicolumn{1}{c}{SSIM$_\text{fg} \uparrow$} & \multicolumn{1}{c}{SSIM$_\text{bg} \uparrow$} & \multicolumn{1}{c}{LPIPS$_\text{fg} \downarrow$} & \multicolumn{1}{c}{LPIPS$_\text{bg} \downarrow$} \\ \cmidrule(r){1-1} \cmidrule(lr){2-2} \cmidrule(lr){3-3} \cmidrule(lr){4-4} \cmidrule(lr){5-5} \cmidrule(lr){6-6} \cmidrule(lr){7-7} \cmidrule(lr){8-8} \cmidrule(lr){9-9} \cmidrule(l){10-10}
\multicolumn{10}{l}{\textit{Fine-tuning Methods}}                                                                                                                                                                                                                                                                                                                                                                                                 \\
LayerDiffuse         & 61.46                                & \underline{0.653}                     & 0.312                                 & 14.78                                         & \textbf{32.76}                                & 0.828                                         & \textbf{0.957}                                & 0.323                                            & \underline{0.039}                                \\ \cmidrule(r){1-1} \cmidrule(lr){2-2} \cmidrule(lr){3-3} \cmidrule(lr){4-4} \cmidrule(lr){5-5} \cmidrule(lr){6-6} \cmidrule(lr){7-7} \cmidrule(lr){8-8} \cmidrule(lr){9-9} \cmidrule(l){10-10}
\multicolumn{10}{l}{\textit{Training-free Methods}}\\                                  
Alfie + inpaiting    & 85.93                                & 0.644                                 & 0.302                                 & 15.32                                         & \underline{27.45}                             & 0.778                                         & \underline{0.947}                             & 0.254                                            & \textbf{0.019}                                   \\
\textbf{Ours}        & \underline{60.53}                    & 0.646                                 & \underline{0.323}                     & \underline{20.46}                             & 25.86                                         & \underline{0.901}                             & 0.895                                         & \underline{0.137}                                & 0.106                                            \\ \cmidrule(r){1-1} \cmidrule(lr){2-2} \cmidrule(lr){3-3} \cmidrule(lr){4-4} \cmidrule(lr){5-5} \cmidrule(lr){6-6} \cmidrule(lr){7-7} \cmidrule(lr){8-8} \cmidrule(lr){9-9} \cmidrule(l){10-10}
\multicolumn{10}{l}{\textit{Application w/ Layout Specification}}\\
\textbf{Ours}        & \textbf{55.59}                       & \textbf{0.655}                        & \textbf{0.329}                        & \textbf{23.82}                                & 23.55                                         & \textbf{0.969}                                & 0.863                                         & \textbf{0.045}                                   & 0.138                                            \\ \bottomrule
\end{tabular}
\label{tab:comparison}
\end{table*}

\section{Experiments}
\subsection{Experimental Setup}

We adopt SDXL~\cite{podell2023sdxl} as LDM for all experiments and generate images at a resolution of $1024\times1024$ using the EulerDiscrete scheduler~\cite{karras2022elucidating} with 50 denoising steps. The guidance scale is set to 7.5 for foreground generation and 5.0 for all other cases. To extract the intermediate latent for seedling noise $L_{\text{fg}}$ and $L_{\text{bg}}$, we set $r_{\text{crop}} = 0.5$, corresponding to the halfway point of the total denoising process. We compare TAUE with representative layer-wise generation baselines, including the fine-tuned method LayerDiffuse~\cite{zhang2024transparent}\footnote{https://github.com/lllyasviel/LayerDiffuse}, and the training-free method Alfie~\cite{quattrini2024alfie}\footnote{https://github.com/aimagelab/Alfie}, combined with a background generation pipeline based on outpainting and inpainting~\cite{zhang2023adding}. Other existing methods were not included in our comparison, as their model weights or datasets are not publicly available, precluding reproducible evaluation.

\subsection{Dataset and Metrics}

We construct a benchmark of 1,770 images filtered from the MS-COCO dataset~\cite{lin2014microsoft}, considering limitations of existing methods in handling multi-object generation and extremely small objects. Specifically, we exclude samples where \texttt{iscrowd} = \texttt{true} or the bounding box area is less than \(0.03\) relative to image size. This ensures each image contains a single, reasonably sized foreground object suitable for fair comparison.
To generate separate prompts for foreground $T_{\text{fg}}$ and background $T_{\text{bg}}$, we employ Phi-3~\cite{abdin2024phi}, a language model-based prompt captioning tool. To evaluate overall image quality, we use Fréchet Inception Distance (FID). For semantic consistency, we compute CLIP-Text (CLIP-T) and CLIP-Image (CLIP-I) similarity scores~\cite{hessel2021clipscore}, which assess alignment with the textual description and visual fidelity, respectively. To further evaluate the disentanglement of foreground and background from complete scenes, we compute PSNR, SSIM, and LPIPS~\cite{zhang2018perceptual} scores between corresponding regions of the composite image and the individually foreground and background layers.

\subsection{Qualitative Result}
Fig.~\ref{fig: result} presents qualitative results of our method and existing methods. Each method outputs foreground, background, and composite images for the same prompt.
Alfie with inpainting, a training-free pipeline, generates the background via outpainting followed by inpainting after foreground generation. However, this often causes foreground features to bleed into the background due to mask misalignment and excessive outpainting. As a result, background images frequently contain residual foreground traces, thereby compromising layer independence. LayerDiffuse, a fine-tuned model, produces better separation and fewer artifacts. However, it still suffers from foreground detail and imperfect semantic harmonization loss. Composite images occasionally exhibit lighting or shadow inconsistencies between layers.

In contrast, our TAUE generates all layers from a shared latent space and attention, ensuring semantic and structural coherence without requiring fine-tuning. The resulting foreground, background, and composite outputs remain visually consistent, with no conflicts or missing elements.

\begin{table*}[t]
\centering
\caption{\textbf{Ablation on high-pass filtering and crop ratio.}
Our default (\emph{50\% + High-pass}) achieves the best holistic quality (lower FID, higher CLIP) while keeping strong reconstruction.
Removing the high-pass slightly improves some reconstruction metrics but visibly hurts perceptual quality (edge softness / occasional duplications).
Extracting the seedling too late (75\%) boosts reconstruction scores but degrades harmonization; too early (25\%) leaves residual noise and lowers text alignment.}
\resizebox{0.95\textwidth}{!}{
\begin{tabular}{@{}lrrrrrrrrr@{}}
\toprule
 & \multicolumn{3}{c}{Overall Quality} & \multicolumn{6}{c}{Layer-wise Reconstruction Quality} \\
\cmidrule(r){2-4}\cmidrule(l){5-10}
Method & FID $\downarrow$ & CLIP-I $\uparrow$ & CLIP-S $\uparrow$ &
PSNR$_\text{fg}\uparrow$ & PSNR$_\text{bg}\uparrow$ &
SSIM$_\text{fg}\uparrow$ & SSIM$_\text{bg}\uparrow$ &
LPIPS$_\text{fg}\downarrow$ & LPIPS$_\text{bg}\downarrow$ \\
\midrule
\textbf{Ours (50\% + High-pass)} & \textbf{55.59} & \textbf{0.655} & \textbf{0.329} & 23.82 & 23.55 & 0.969 & 0.863 & 0.045 & 0.138 \\
50\% \, w/o High-pass            & 55.79 & 0.654 & 0.328 & 23.92 & 23.59 & 0.970 & 0.862 & 0.045 & 0.139 \\
75\% (late)                      & 56.48 & 0.653 & 0.328 & \textbf{24.33} & \textbf{25.02} & \textbf{0.974} & \textbf{0.904} & \textbf{0.041} & \textbf{0.091} \\
25\% (early)                     & 55.70 & 0.640 & 0.321 & 21.12 & 19.70 & 0.953 & 0.750 & 0.059 & 0.284 \\
\bottomrule
\end{tabular}
}
\label{tab:ablation_highpass_cropratio}
\end{table*}

\begin{figure*}[t]
\centering
\includegraphics[width=0.95\textwidth]{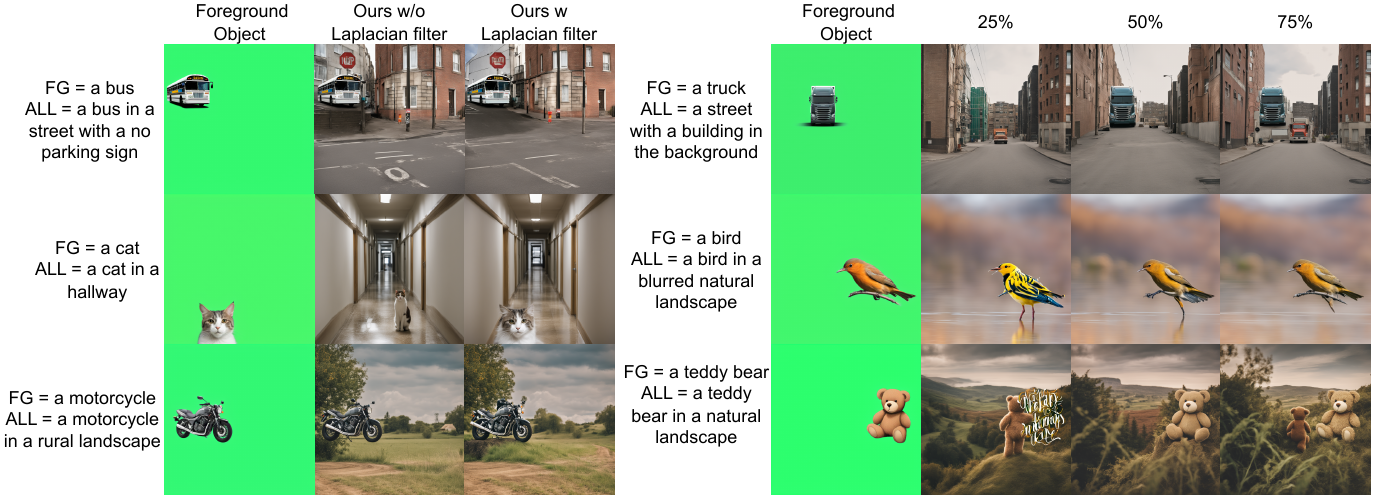}
\caption{
\textbf{Ablation on high-pass filtering and crop ratio.}
Left: removing the Laplacian filter leads to faded or floating objects, while applying it preserves high-frequency details and yields sharper, coherent composites. 
Right: varying the crop ratio (25\%, 50\%, 75\%) controls when the seedling latent is extracted—25\% fail to reconstruct, 75\% overfit to the foreground, and 50\% achieves the best balance with consistent foreground–background integration.
}
\label{fig:ablation_combined}
\end{figure*}

\subsection{Quantitative Result}
As shown in Tab.~\ref{tab:comparison}, TAUE achieves the best image quality among training-free methods across all metrics. Our method outperforms the fine-tuned LayerDiffuse in FID and CLIP-S, indicating superior visual fidelity and stronger alignment with the text. While LayerDiffuse slightly leads in CLIP-I, TAUE remains highly competitive overall.
For layer-wise reconstruction, TAUE achieves the highest foreground accuracy across all metrics (PSNR, SSIM, LPIPS), demonstrating that the transplanted seedling noise effectively preserves object details. Background reconstruction scores are slightly lower than those of Alfie with inpainting and LayerDiffuse, both of which benefit from reusing unmasked background pixels—an advantage that artificially boosts their scores. 

In contrast, TAUE performs denoising entirely from scratch, relying solely on intermediate seedling noise without reusing any pixel values. This increases the difficulty of the task, making it more challenging, and reduces the likelihood of regenerating the foreground object in precisely the same position and structure. 
Nevertheless, TAUE maintains a strong background quality and overall consistency.

\begin{figure*}[t]
\centering
\includegraphics[width=0.95\textwidth]{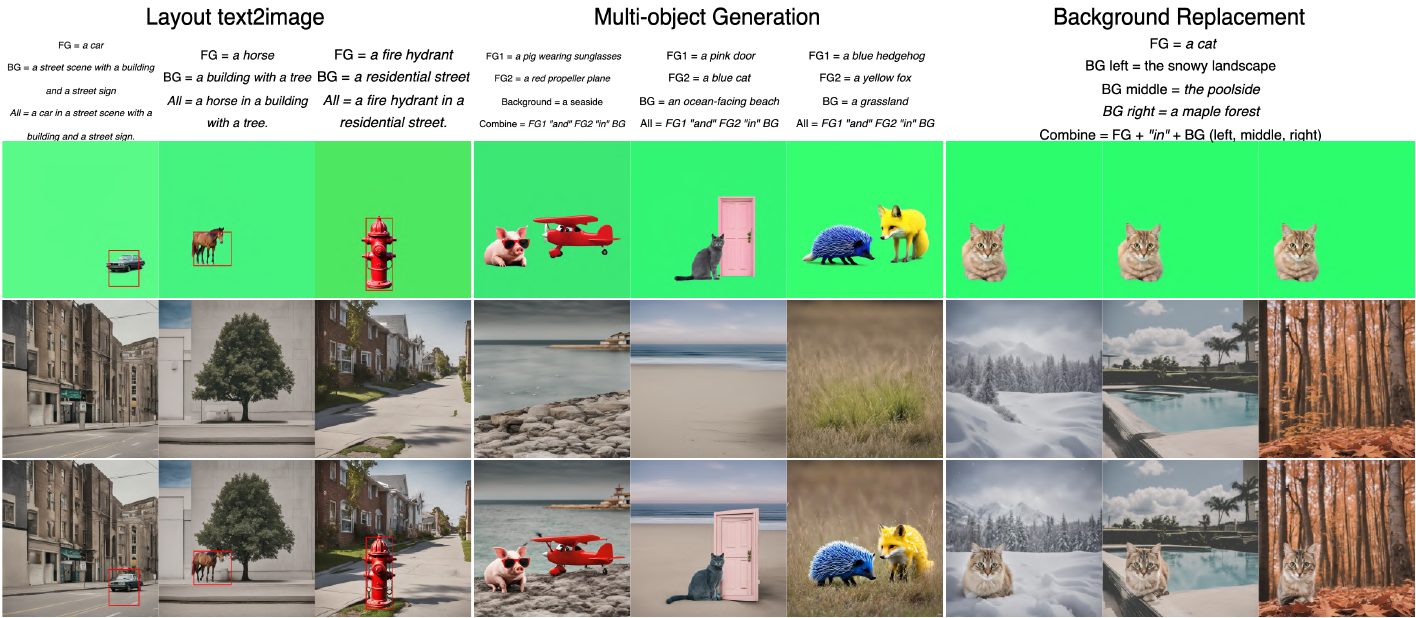} 
\caption{
    Applications of TAUE. TAUE enables (a) \textit{Layout and Size Control} by injecting bounding box constraints to specify the position and scale of the foreground object; (b) \textit{Disentangled Multi-Object Generation} by transplanting seedling noise to multiple spatial locations, allowing compositionally coherent and semantically independent objects; and (c) \textit{Background Replacement} by regenerating backgrounds while preserving the original foreground structure. 
  }
\label{fig: application}
\end{figure*}

\subsection{Ablation Study}
As shown in Tab.~\ref{tab:ablation_highpass_cropratio} and Fig.~\ref{fig:ablation_combined}, we analyze the effect of the Laplacian high-pass filter and the crop ratio used to extract the seedling latent. Further ablation studies are provided in the supplementary material.

\noindent
\textbf{Laplacian high-pass filter}
Removing the high-pass filter slightly improves reconstruction metrics but degrades perceptual and image quality. 
The absence of the filter prevents the model from retaining foreground-aware structural information, resulting in artifacts such as faded facial regions in the cat example or floating motorcycles detached from the ground. 
In contrast, applying the filter preserves high-frequency cues in the transplanted latent, producing sharper edges and visually coherent composites.

\noindent
\textbf{Choice of Crop Ratio}
We also evaluate different crop ratios (25\%, 50\%, 75\%) for latent extraction. 
Early extraction (25\%) fails to capture sufficient foreground structure, often producing incorrect object shapes, whereas late extraction (75\%) overemphasizes reconstruction fidelity, yielding floating or scene-inconsistent objects similar to the original foreground. 
The mid-point setting (50\%) provides the best trade-off between structural preservation and generative flexibility, achieving the most coherent composites with balanced foreground–background integration.

\section{Applications}
\label{sec:applications}

We demonstrate TAUE's utility in three applications: layout and size control, multi-object generation with disentangled attributes, and background replacement without fine-tuning and special data in Fig.~\ref{fig: application} and Tab~\ref{tab:comparison}.

\noindent\textbf{Layout and Size Control~~}
Existing layer-wise image generation models often produce centrally aligned, uniformly sized objects due to limited spatial conditioning~\cite{zhang2024transparent, quattrini2024alfie}. Controlling object scale via prompts or layout models often lacks consistency and flexibility. TAUE addresses these limitations by injecting user-defined bounding boxes, which specify the foreground object's position and size. It guides seedling noise transplantation and denoising to generate semantically coherent content within the desired region.

As shown in Tab.~\ref{tab:comparison}, the layout-aware TAUE improves FID, CLIP-I, CLIP-S, and foreground reconstruction (PSNR, SSIM, LPIPS). Explicit size control also enhances alignment with object semantics, yielding more accurate object scales. While background reconstruction is slightly lower due to full denoising instead of pixel reuse, this trade-off improves harmonization and realism.

\noindent\textbf{Disentangled Multi-Object Generation~~}
Text-to-image models often suffer from attribute entanglement when generating multiple objects in a single prompt like incorrect color or shape assignments~\cite{chefer2023attend}. Prior works~\cite{zhang2024transparent} mitigate this by sequential generation and compositing, but incur high inference cost and blending artifacts.

TAUE introduces an efficient alternative by transplanting seedling noise to multiple spatial locations in the latent space. Each transplanted latent preserves its object semantics, enabling simultaneous generation of multiple, disentangled objects in a single denoising process. (Fig.~\ref{fig: application},)

\noindent\textbf{Background Replacement~~}
Conventional diffusion models regenerate the full image to modify the background, which risks altering the foreground due to entangled representations. TAUE resolves this by decoupling foreground and background generation via latent transplantation.
Once the foreground object is generated, its seedling noise can be retained and reused to synthesize a new background independently. This ensures a consistent foreground appearance and layout. Additionally, adjusting transplantation coordinates allows repositioning of the object across backgrounds, enabling interactive layout-aware editing.

Unlike existing methods that rely on handcrafted prompts or retraining, TAUE enables seamless background edits. This enables rapid iteration and fine-grained control in creative workflows, such as UI design and ad generation.
\section{Limitation}
In cases requiring high-fidelity foreground preservation, e.g., when the exact shape, color, or pixel-level structure of the foreground must remain unchanged, TAUE may underperform compared to inpainting-based methods that modify the background while preserving the foreground. Future work should explore ways to control this behavior, such as selectively freezing foreground features during compositing or introducing constraints to balance semantic adaptation and structural preservation. Addressing this trade-off between harmonization and fidelity will be crucial to expanding TAUE's applicability in precision-critical tasks.
\section{Conclusion}

We introduced TAUE, a training-free framework that enables layer-wise image generation directly within the latent space of diffusion models, allowing structural and semantic information to propagate across layers with minimal computational overhead. 
Experimental results show that TAUE matches or surpasses fine-tuned models in both qualitative and quantitative evaluations. 
Beyond quantitative performance, our framework supports controllable and practical applications such as layout-guided synthesis, multi-object composition, and background replacement, bridging the gap between generative research and real-world creative workflows. 
We believe TAUE represents a step toward modular, accessible, and layer-aware image generation, opening new directions for diffusion-based content creation and editing.

{
    \small
    \bibliographystyle{ieeenat_fullname}
    \bibliography{main}
}


\end{document}